\documentclass{article}
\usepackage{stywhispers,amsmath,epsfig}
\usepackage[font=small, skip=0pt]{caption}
\usepackage{amsfonts} 
\usepackage{float}
\usepackage{xcolor}
\usepackage{enumitem}

\usepackage{amsmath}
\title{Spectral-Enhanced Transformers: Leveraging Large-Scale Pretrained Models for Hyperspectral Object Tracking}
\name{Shaheer Mohamed$^{1,2}$, Tharindu Fernando$^{1}$, Sridha Sridharan$^1$, Peyman Moghadam$^{2,1}$, Clinton Fookes$^1$ }
\address{$^1$Signal Processing, Artificial Intelligence and Vision Technologies, Queensland University of Technology, \\
Brisbane, Australia \\
$^2$Robotics and Autonomous Systems, Data61, CSIRO, Brisbane, QLD, Australia}

\begin{document}
%
\maketitle
%
\renewcommand{\thefootnote}{}
\footnotetext{This research was supported by an Australian Research Council (ARC) Discovery Grant DP200101942.}
\renewcommand{\thefootnote}{\arabic{footnote}}

\begin{abstract}
Hyperspectral object tracking using snapshot mosaic cameras is emerging as it provides enhanced spectral information alongside spatial data, contributing to a more comprehensive understanding of material properties. Using transformers, which have consistently outperformed convolutional neural networks (CNNs) in learning better feature representations, would be expected to be effective for Hyperspectral object tracking. However, training large transformers necessitates extensive datasets and prolonged training periods. This is particularly critical for complex tasks like object tracking, and the scarcity of large datasets in the hyperspectral domain acts as a bottleneck in achieving the full potential of powerful transformer models. This paper proposes an effective methodology that adapts large pretrained transformer-based foundation models for hyperspectral object tracking. We propose an adaptive, learnable spatial-spectral token fusion module that can be extended to any transformer-based backbone for learning inherent spatial-spectral features in hyperspectral data. Furthermore, our model incorporates a cross-modality training pipeline that facilitates effective learning across hyperspectral datasets collected with different sensor modalities. This enables the extraction of complementary knowledge from additional modalities, whether or not they are present during testing. Our proposed model also achieves superior performance with minimal training iterations.

\end{abstract}
\section{Introduction}
\label{sec:intro}

\begin{figure}[t]
    \centering
    \includegraphics[scale=0.6]{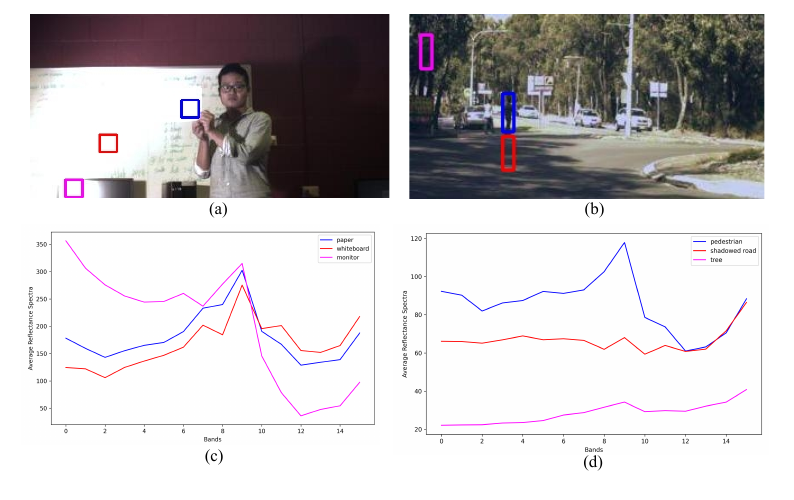}
    \caption{Example of different objects with similar visual cues but distinct spectral curves.}
    \label{fig:intro}
\end{figure}

Hyperspectral object tracking is gaining significant attention in modern computer vision due to its ability to perceive beyond the visual spectrum \cite{MMF-Net, TBR-Net}. Hyperspectral images capture both spatial and spectral information, with the spectra reflecting the physical and material properties of objects. This advanced perception is particularly valuable in applications where visual ambiguity can lead to misinterpretations. As illustrated in Fig. \ref{fig:intro}, in complex scenes where distinguishing between objects is challenging when only visual information is vailable but the spectral information offers clear distinguishing features. When considering hyperspectral images captured by snapshot cameras, these systems acquire entire scenes at once using a mosaic pattern, resulting in fewer bands compared to line or point scanning cameras, which can capture hundreds of bands. The advent of snapshot cameras has enabled real-time hyperspectral image acquisition, thereby facilitating complex tasks like object tracking. Consequently, new feature-learning approaches should be adopted to achieve optimal performance, leveraging both spatial and complementary spectral information.

In modern deep learning, transformer-based networks have revolutionized natural language processing. Recently, they have increasingly become popular in the computer vision domain, replacing CNNs in many tasks \cite{lin2022swintrack, arnab2021vivit}. Additionally, recent foundation models trained on very large datasets over extended periods show promise in learning generalizable features. However, training such large foundation models with hyperspectral data is nearly impossible due to limited data availability. Therefore, we focus on how we can effectively adapt pretrained transformers for hyperspectral applications by enhancing them with spectral features. 

To bridge the gap and utilize pretrained transformer-based foundation models for snapshot hyperspectral data, we propose a spectral-guided transformer model that utilizes pretrained weights for RGB images. Recognizing that pretrained models on RGB images are proficient at extracting spatial features, we employ false-color images generated from the hyperspectral data for learning spatial features. Simultaneously, we incorporate complementary spectral embeddings to enrich the model with the spectral information inherent in hyperspectral images. Here, we adaptively learn the input embeddings of the transformer to benefit from spectral features while maintaining the efficacy and generalization capabilities of the pretrained spatial features. This approach enables us to effectively apply large pretrained models to hyperspectral data, achieving superior performance even with limited training data. Also our proposed model is based on a Siamese framework and can be trained and tested across varying number hyperspectral modalities such that it learns complementary cross-modal information. As such, our framework allows the extraction of complementary knowledge from additional modalities despite those modalities are not present during the test time. Our contributions are listed below.
\begin{itemize}[noitemsep, topsep=0pt, leftmargin=*]
  \item We propose a fully transformer-based pipeline for hyperspectral object tracking that effectively leverages large-scale pretrained weights from image-based models, enabling convergence within a few epochs.
  \item We introduce an adaptive, learnable spatial-spectral token fusion model that efficiently integrates spatial and spectral features in a complementary manner. This module can be extended to any transformer-based backbone. 
  \item Our method supports cross-modality training across multiple hyperspectral datasets with varying bands. It learns modality-invariant features and demonstrates robust performance when evaluated on single modalities.
\end{itemize}
\section{Related Works}
\label{sec:format}
Recent advancements in the RGB image domain have demonstrated the effectiveness of fully transformer-based pipelines for object tracking \cite{lin2022swintrack, cui2024mixformerv2}. These works clearly show that having a pretrained transformer-based backbone plays a crucial role in learning better feature representations for object tracking. In contrast, most previous methods for hyperspectral object tracking rely on CNN-based Siamese networks or hybrid networks that utilize CNNs for feature extraction \cite{MMF-Net, TransDAT}, and self-attention mechanisms of transformers are only used for feature fusion. Consequently, these approaches fail to leverage the full potential of transformers as backbones.

However, limited works that utilize transformers for hyperspectral object tracking \cite{TBR-Net, SPIRIT}. These methods often employ spectral dimensionality reduction techniques to reduce the number of bands to three, allowing pretrained RGB image weights of the transformer to be directly adopted. However, such methods do not fully exploit the rich spectral information inherent in hyperspectral data, and their performance primarily rely upon the effectiveness of the dimensionality reduction process.

In our approach, we address this limitation by inputting the full spectral data into the network, allowing it to adaptively learn salient spectral and spatial features in a learnable manner. This enables us to fully leverage the capabilities of transformers without sacrificing the spectral richness of hyperspectral images. Moreover, our framework adaptively learns across varying numbers of spectral bands, ensuring superior test performance even when some bands are missing during testing. 
\vspace{-0.1cm}
\section{Proposed Method}
\label{sec:pagestyle}
\begin{figure*}[!t]
    \centering
    \includegraphics[scale=0.90]{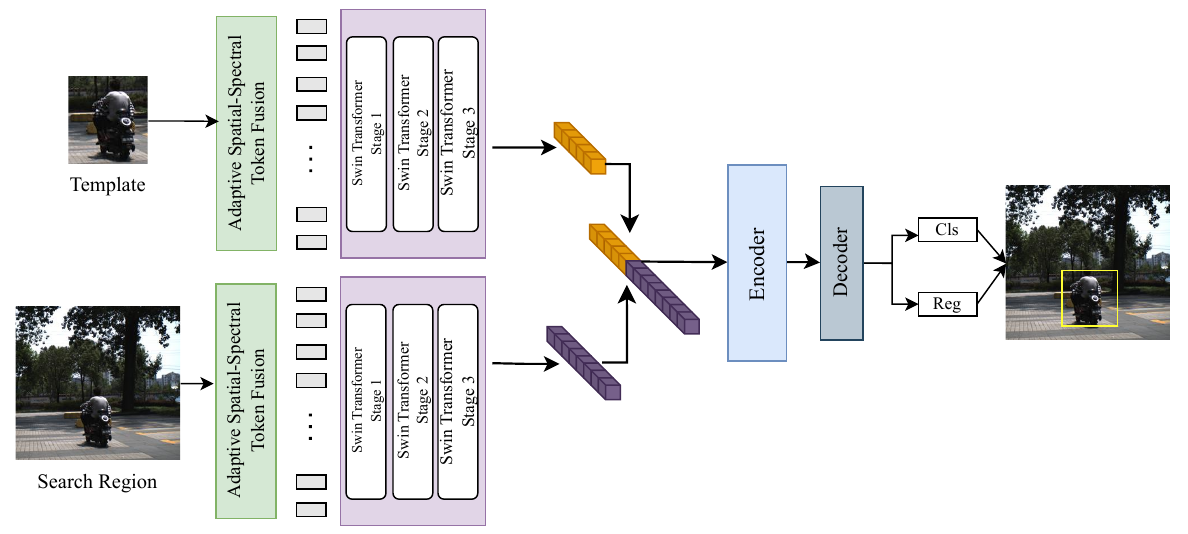}
    \caption{An overview of the proposed architecture for tracking. Adaptive, learnable spatial-spectral token fusion merges spatial and spectral features, which are then processed through a Swin Transformer backbone. A transformer-based encoder-decoder is employed for further feature fusion, followed by a prediction head.}
    \label{fig:overview}
\end{figure*}

In this section we describe the proposed method. We propose a fully transformer based pipeline for hyperspectral object tracking that utilizes large-scale pre-trained weights. Specifically, we capture our inspiration from SwinTrack \cite{lin2022swintrack} siamese tracking pipeline \cite{lin2022swintrack} due to its superior performance in the RGB image domain and extend it to extract knowledge from hyperspectral modalities. The overall network architecture is shown in Fig \ref{fig:overview} and details of its main components are illustrated in the following subsections. 

\subsection{Adaptive Spatial-Spectral Token Fusion}
\vspace{-0.1cm}
In transformer-based networks, the initial step is to tokenize the input image. The input false color image $X_{fc} \in \mathbb{R}^{3 \times H \times W}$ of size $H \times W$ and 3 bands and the hyperspectral image $X_{hsi} \in \mathbb{R}^{B \times H \times W}$, with $B$ number of bands, are divided into non-overlapping patches of size $16 \times 16$. Each patch is defined as $y_{fc} \in \mathbb{R}^{3 \times 16 \times 16}$ and $y_{hsi} \in \mathbb{R}^{B \times 16 \times 16}$. Next, the patches are flattened and projected into an embedding of size $d$ using linear projection matrices $\mathbf{E}_{fc} $ and $\mathbf{E}_{hsi} \in \mathbb{R}^{d \times M}$ as shown in Eq. \ref{eq:fc-proj} and \ref{eq:hsi-proj}, where $M = H/16 \times W/16$ is the total number of patches.

\begin{equation}
    \mathrm{z}_{(fc,i)}=\mathbf{E}_{(fc)} . y_{(fc,i)} 
    \quad i=1,...,M,
    \label{eq:fc-proj}
\end{equation}
and
\begin{equation}
    \mathrm{z}_{(hsi,i)}=\mathbf{E}_{(hsi)} . y_{(hsi,i)}
    \quad i=1,...,M.
    \label{eq:hsi-proj}
\end{equation}

\begin{figure}[!t]
    \centering
    \includegraphics[scale=0.8]{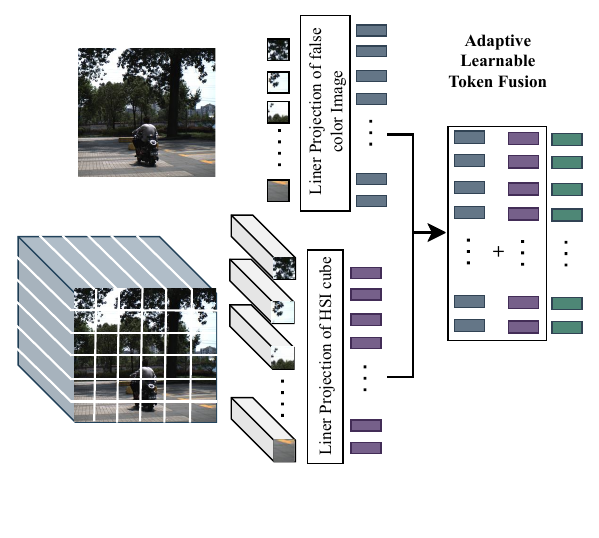}
    \caption{Proposed Adaptive Spatial-Spectral Token Fusion Module}
    \label{fig:ssp}
\end{figure}

This process of converting patches into tokens is called patch embedding or tokenization. This layer learns the inherent structure of the input using a single linear layer before the tokens are passed into the transformer block. Therefore, the entire transformer learns the features and dependencies using these input tokens, and it's vital to tokenize them properly to fully exploit the potential of pretrained transformer networks. Since snapshot hyperspectral cameras possess spatial and spectral information, we use two linear projection layers to tokenize them separately to learn salient features in both dimensions. Next, since spatial and spectral information are complementary to each other, meaning that in some patches, spatial information could be more salient while spectral data might be noisy, and vice versa, as in Eq. \ref{eq:fusion}, we propose an adaptive spatial-spectral token fusion using a learnable parameter, $\alpha$. This allows the proposed model to effectively leverage the strengths of both modalities into object tracking. The overall architecture of the adaptive spatial-spectral token fusion module is illustrated in Fig. \ref{fig:ssp}.  
\begin{equation}
    \mathrm{z}_{(i)} = \alpha_i \times z_{(fc, i)} + (1-\alpha_i) \times z_{(hsi, i)}
    \label{eq:fusion}
\end{equation}

Although we implemented our proposed method using a Swin Transformer backbone, via adopting and fusing spectral information during the patch embedding phase, our approach remains flexible and adaptable any transformer based backbone.

\subsection{Adapting Pretrained Weights and Training Across Modalities }\label{sec:pre-trained_weights}
\vspace{-0.1cm}
When loading pretrained weights to the SwinTrack \cite{lin2022swintrack} framework, we only need to handle the patch embedding weights, since no other part in this architecture is changed. For the patch embedding weights of false color images, we use the same weights from the RGB model. For Hyperspectral patch embedding weights we inflate the RGB weights as described in \cite{arnab2021vivit}. 

Additionally, since hyperspectral data are limited, we perform training across modalities. Using hyperspectral images captured with different sensors, we train them together, allowing the model to learn sequentially across these numerous modalities allowing it to capture complementary information across distinct modalities. However, this requires handling scenarios with differences in the number of bands and modalities during patch embedding stage, where we use zero-padding for unavailable bands. In the testing phase, we test individual modalities separately. As such, this approach has similarities to cross-modality training and uni-modal testing methodology.

\section{Experiments}
\label{sec:typestyle}
In this section, we describe our experimental setup, provide
descriptions of the datasets, and outline the evaluation metrics used. We also present our results, including comparisons with state-of-the-art (SOTA) methods.
\vspace{-0.15cm}
\subsection{Experimental Setup}
\vspace{-0.1cm}
We use HOT2020 and HOT2024, two public datasets released by the Hyperspectral Object Tracking challenge in 2020 and 2024 respectively. HOT2020 includes 40 training and 35 testing videos, while HOT2024 expands to 182 training and 89 validation sequences captured across VIS, NIR, and RedNIR spectra (16, 25, and 15 bands). Both datasets includes the false color versions as well.

Our model is implemented using PyTorch and trained on two NVIDIA A100 GPUs. We load the tiny version of the Swin Transformer weights with an embedding size of 384 as discussed in Sec. \ref{sec:pre-trained_weights}. The model is trained for 3 and 5 epochs on the HOT2020 and HOT2024 datasets.Precision plots and success plots are used for evaluation. We compute the Area Under the Curve (AUC) of the success plots and measure distance precision with a threshold of 20 pixels (DP\_20) as in \cite{MMF-Net}.

\subsection{Results and Discussion}
\vspace{-0.1cm}
\subsubsection{Comparison on HO2020 Dataset}
\vspace{-0.1cm}
We first compare our approach with current SOTA hyperspectral and visual trackers on the HOT2020 dataset, and the quantitative results are shown in Table \ref{tab:HOT2020}. Among the hyperspectral trackers, BAE-Net \cite{BAE-Net} adopts a band attention-aware ensemble network to generate false-color images; SiamBAG \cite{SiamBAG} proposes a band regrouping Siamese network for generating three-channel images that utilize RGB trackers to enhance hyperspectral tracking performance; and SST-Net \cite{SST-Net} introduces a spatial-spectral-temporal attention network for learning salient features. Additionally, we compare our model's performance with recent RGB trackers such as TransT \cite{TransT}, SiamGAT \cite{SiamGAT}, SimTrack \cite{SimTrack}, OSTrack \cite{OSTrack}, and SwinTrack \cite{lin2022swintrack}.

Our proposed model achieves the highest AUC score of \textbf{0.647} and the second-best DP\_20 score of \textbf{0.889}, outperforming both hyperspectral and visual trackers. Notably, our model attains this high performance with only three epochs of training, demonstrating its effectiveness in leveraging large pretrained models. Moreover, the performance improvement compared to SwinTrack, which follows a similar tracking pipeline without the proposed adaptive spatial-spectral module, further validates that adaptively fusing spectral information is beneficial.
\vspace{-0.45cm}
\subsubsection{Comparison on HO2024 Dataset}
\vspace{-0.1cm}
To demonstrate the flexibility of our approach to HSI data with a variable number of modalities and bands we also evaluate our approach on the HOT2024 dataset, which includes hyperspectral images captured by three different sensors. Table \ref{tab:HOT2024} presents the quantitative performance for each sensor modality as well as the average performance across all modalities. Among the competing methods, MMF-Net \cite{MMF-Net} introduces a material-guided multi-view fusion network that integrates material information with false-color images. TransDAT \cite{TransDAT} is a domain-adaptive transformer-based network tailored for hyperspectral tracking. SPIRIT \cite{SPIRIT} proposes an end-to-end spectral-aware network with a dynamic template, leveraging RGB pretrained weights for improved performance. SEE-Net \cite{SEE-Net} employs a deep ensemble network with band regrouping, utilizing a spectral-self-awareness module to enhance feature extraction.

When comparing these methods with our proposed model, we achieve the highest accuracy across all modalities with clear margins, attaining an AUC score of 0.506 on VIS, 0.759 on NIR, and 0.465 on RedNIR. The significant performance gap between our model and the second-best performers, particularly in the NIR and RedNIR modalities further demonstrates that our cross-modality training and uni-model testing effectively learns modality-invariant features.

\begin{table*}[]
\centering
\caption{Overall Performance (AUC and DP\_20) Comparison of Hyperspectral and Visual Trackers on HOT2020 Dataset. The best results are bold and second best results are underlined.}
\label{tab:HOT2020}
\resizebox{\textwidth}{!}{\begin{tabular}{c||c c c c c c c c c c}
\hline 
Method & Ours & BAE-Net \cite{BAE-Net} & SiamBAG \cite{SiamBAG} & SST-Net \cite{SST-Net} & TransT \cite{TransT} & SiamGAT \cite{SiamGAT} & SimTrack \cite{SimTrack} & OSTrack \cite{OSTrack} & SwinTrack \cite{lin2022swintrack}  \\
\hline
AUC & \textbf{0.647} & 0.606 & 0.622 & 0.623 & 0.633 & 0.581 & 0.600 & 0.557 & \underline{0.637} & \\
DP\_20  & \underline{0.889} & 0.878 & 0.877 &  \textbf{0.916} &0.87 & 0.827 & 0.845 & 0.816 & 0.866 & \\
\hline 
\end{tabular}}
\end{table*}



\begin{table}[]
\centering
\caption{Overall Performance (AUC and DP\_20) Comparison of Hyperspectral and Visual Trackers on HOT2024 Dataset. The best results are bold and second best results are underlined.}
\label{tab:HOT2024}
\resizebox{\columnwidth}{!}{\begin{tabular}{c|c|ccccc}
\hline
 & Method & Ours & MMF-Net  & TransDAT  & SPIRIT & SEE-Net   \\
\hline
VIS & AUC & \textbf{0.506} &  \underline{0.482} & 0.397 & 0.319 & 0.396   \\
& DP\_20 & \textbf{0.678} & \underline{0.645} & 0.524 & 0.409 & 0.560 \\
\hline
NIR & AUC & \textbf{0.759} & \underline{0.701} & 0.587 & 0.656 & 0.509   \\
& DP\_20 & \textbf{0.915} & \underline{0.876} & 0.754 & 0.824 & 0.769  \\
\hline
RedNIR & AUC & \textbf{0.465} & 0.388 & \underline{0.423} & 0.377 & 0.383   \\
& DP\_20 & \textbf{0.632} & 0.521 & \underline{0.547} & 0.516 & 0.521  \\
\hline 
Total & AUC & \textbf{0.564} & \underline{0.527} & 0.453 & 0.417 & 0.426   \\
& DP\_20 & \textbf{0.730} & \underline{0.683} & 0.587 & 0.534 & 0.607 \\
\hline 

\end{tabular}}
\end{table}

\begin{figure}[htb]

%
\begin{minipage}[b]{.48\linewidth}
  \centering
  \includegraphics[width=\linewidth]{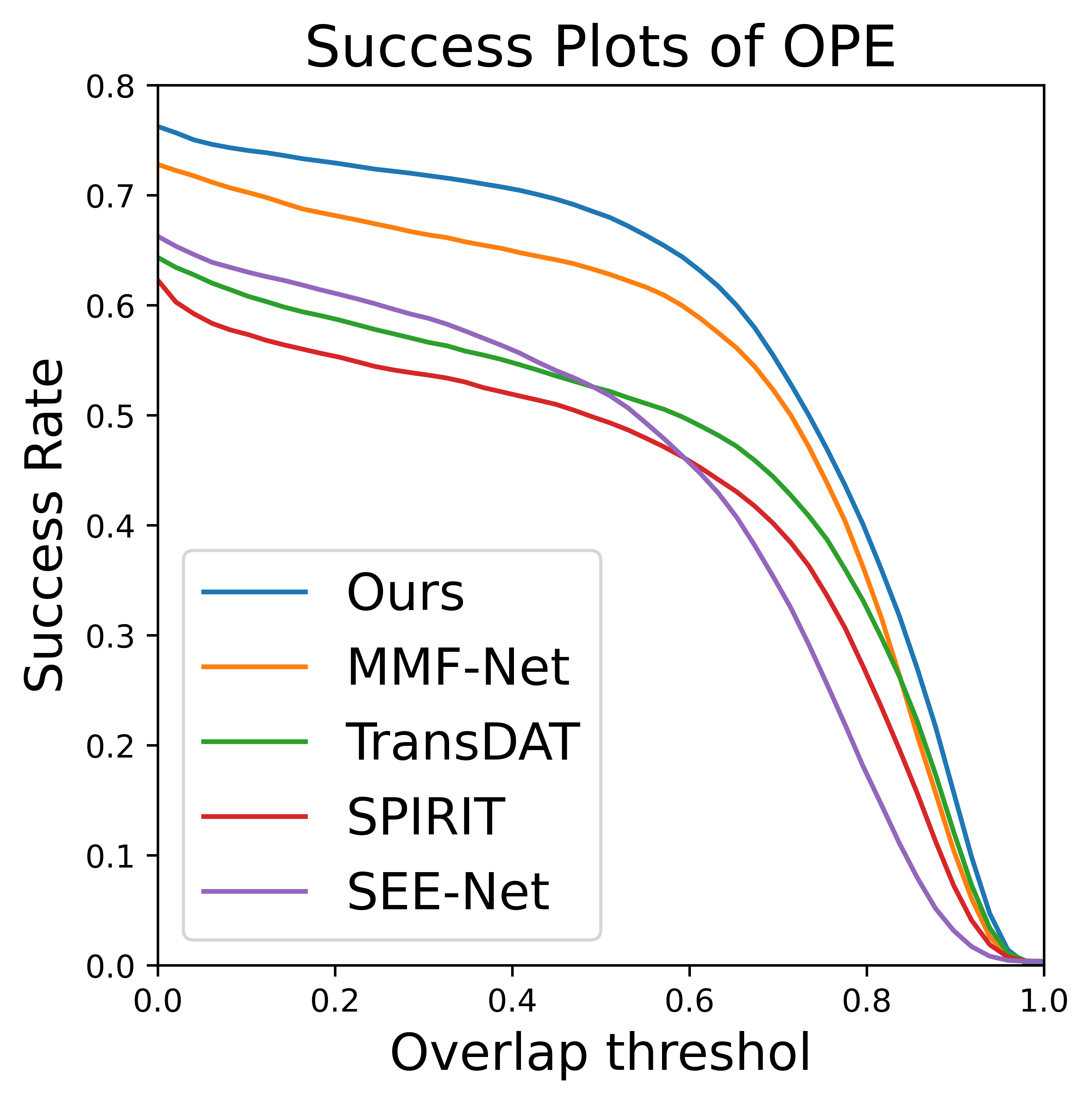} 
  \vspace{0.1cm}
  \centerline{(a) Success plot}\medskip
\end{minipage}
\hfill
\begin{minipage}[b]{.48\linewidth}
  \centering
  \includegraphics[width=\linewidth]{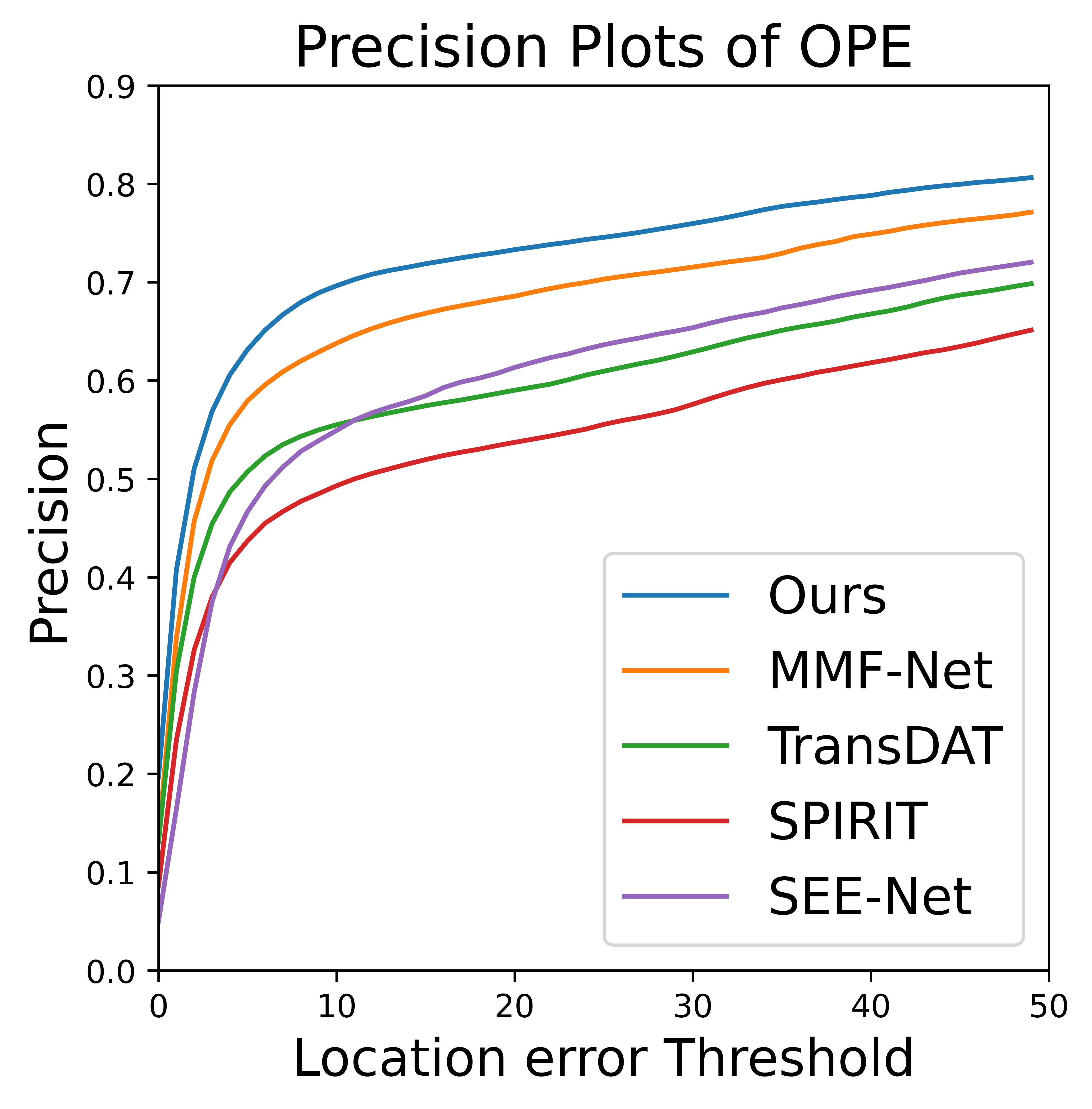} 
  \vspace{0.1cm}
  \centerline{(b) Precision plot}\medskip
\end{minipage}
\caption{Success and Precision plots of the compared trackers on the HOT2024 dataset across all sequences.}
\label{fig:res}
\end{figure}




\vspace{-0.1cm}
\section{Conclusion}
\label{sec:page}

In this paper, we introduce a fully transformer-based tracking pipeline for snapshot hyperspectral object tracking, leveraging large-scale pretrained models on RGB data. We introduce an adaptive spatial-spectral token fusion module that learns to integrate spectral features with spatial features, allowing for the extraction of salient information from both dimensions. Although we employ a Swin Transformer based backbone, our fusion module is compatible with any transformer-based backbone. Additionally, when hyperspectral data from multiple sensor modalities are available, our method enables cross-modality training, allowing the model to learn modality-invariant features. Our results, evaluated on both the HOT2020 and HOT2024 datasets, show that the proposed method efficiently leverages pretrained weights from large models and successfully learns salient spatial-spectral features with only a few training epochs, achieving commendable results. 

\bibliographystyle{IEEEbib}
\bibliography{strings,refs}

\end{document}